\documentclass[11pt,a4paper]{article}
\usepackage[hyperref]{acl2020}
\usepackage{times}
\usepackage{latexsym}

\usepackage{url}

\aclfinalcopy %
\def\aclpaperid{2049}
\usepackage{url}
\usepackage{subcaption} 
\usepackage{booktabs} %
\usepackage{enumitem}
\usepackage{graphicx}
\usepackage{bbm}
\usepackage{amsmath,amsfonts,bm}
\usepackage[T1]{fontenc}
\usepackage[utf8]{inputenc}
\usepackage{multirow}
\usepackage{algorithm}
\usepackage{algpseudocode}
\author{Giannis Karamanolakis\thanks{\hskip 5pt Work performed during internship at Amazon.}\\
  Columbia University \\ New York, NY 10027, USA \\
  \texttt{gkaraman@cs.columbia.edu} \\\And
  Jun Ma, Xin Luna Dong \\
  Amazon.com \\ Seattle, WA 98109, USA \\
  \texttt{\{junmaa, lunadong\}@amazon.com} \\}

\setlist[enumerate,1]{leftmargin=1.2em,labelindent=0em,itemsep=0pt,labelsep*=0.5em}

\title{TXtract: Taxonomy-Aware Knowledge Extraction\\ for Thousands of Product Categories}

\date{}

\begin{document}
\maketitle
\begin{abstract}
Extracting structured knowledge from product profiles is crucial for various applications in e-Commerce.
State-of-the-art approaches for knowledge extraction were each designed for a single category of product, and thus do not apply to real-life e-Commerce scenarios, which often contain thousands of diverse categories. This paper proposes TXtract, a taxonomy-aware knowledge extraction model that applies to thousands of product categories organized in a hierarchical taxonomy. Through category conditional self-attention and multi-task learning, our approach is both scalable, as it trains a single model for thousands of categories, and effective, as it extracts category-specific attribute values. Experiments on products from a taxonomy with 4,000 categories show that TXtract outperforms state-of-the-art approaches by up to 10\% in F1 and 15\% in coverage across \emph{all} categories.
\end{abstract}

\section{Introduction}

\begin{figure}[t]
\centering
\includegraphics[height = 5.8cm, width = 7.8cm]{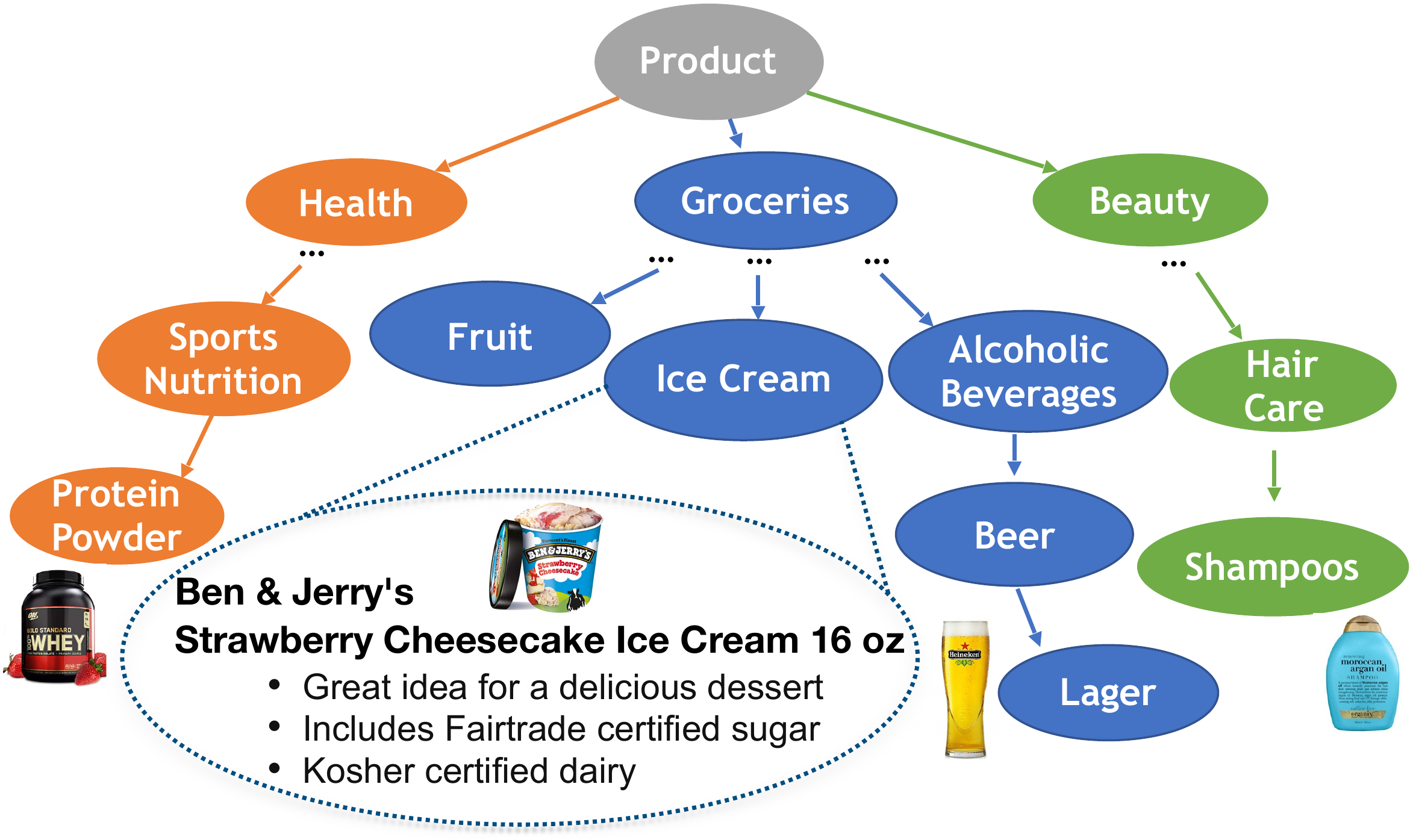}
	\caption{A hierarchical taxonomy with various product categories and the public webpage of a product assigned to ``Ice Cream'' category.}
\label{fig:amazon_product_example}
\end{figure}

Real-world e-Commerce platforms contain billions of products from thousands of different categories, organized in hierarchical taxonomies (see Figure~\ref{fig:amazon_product_example}).
Knowledge about products can be represented in structured form as a catalog of product attributes (e.g., \emph{flavor}) and their values (e.g., ``strawberry''). 
Understanding precise values of product attributes is crucial for many applications including product search, recommendation, and question answering. 
However, structured attributes in product catalogs are often sparse, leading to unsatisfactory search results and various kinds of defects. Thus, it is invaluable if such structured information can be extracted from product profiles such as product titles and descriptions.
Consider for instance the ``Ice Cream'' product of Figure~\ref{fig:amazon_product_example}.
The corresponding title can potentially be used to extract values for attributes, such as ``Ben \& Jerry's'' for \emph{brand}, ``Strawberry Cheesecake'' for \emph{flavor}, and ``16 oz'' for \emph{capacity}.

State-of-the-art approaches for attribute value extraction~\cite{zheng2018opentag,xu2019scaling,rezk2019accurate} have employed deep learning to capture features of product attributes effectively for the extraction purpose. %
However, they are all designed without considering the product categories and thus cannot effectively capture the diversity of categories across the product taxonomy. 
Categories can be substantially different in terms of applicable attributes (e.g., a ``Camera'' product should not have \emph{flavor}), attribute values (e.g., ``Vitamin'' products may have ``fruit'' \emph{flavor} but ``Banana'' products should not) and more generally, text patterns used to describe the attribute values (e.g., the phrase ``infused with'' is commonly followed by a \emph{scent} value such as ``lavender'' in ``Hair Care'' products but not in ``Mattresses'' products).

In this paper, we consider attribute value extraction for real-world hierarchical taxonomies with thousands of product categories, where directly applying previous approaches presents limitations.
On the one extreme, ignoring the hierarchical structure of categories in the taxonomy and assuming a single ``flat'' category for all products does not capture category-specific characteristics and, as we will show in Section~\ref{s:experiments}, is not effective.
On the other extreme, training a separate deep neural network for each category in the product taxonomy is prohibitively expensive, and can suffer from lack of training data on small categories.

To address the limitations of previous approaches under this challenging setting, we propose a framework for \emph{category-specific} attribute value extraction that is both efficient and effective.
Our deep neural network, TXtract, is \emph{taxonomy-aware}: it leverages the hierarchical taxonomy of product categories and extracts attribute values for a product conditional to its category, such that it automatically associates categories with specific attributes, valid attribute values, and category-specific text patterns.
TXtract is trained on all categories in parallel and thus can be applied even on small categories with limited labels. 

The key question we need to answer is {\em how to condition deep sequence models on product categories}. Our experiments suggest that following previous work to append category-specific artificial tokens to the input sequence, or concatenate category embeddings to hidden neural network layers is not adequate. %
There are two key ideas behind our solution. First, we use the category information as context to generate category-specific token embeddings via conditional self-attention. Second, we conduct multi-task training by meanwhile predicting product category from profile texts; this allows us to get token embeddings that are discriminative of the product categories and further improve attribute extraction. Multi-task training also makes our extraction model more robust towards wrong category assignment, which occurs often in real e-Commerce websites.\footnote{Examples: (1) an ethernet cable assigned under the ``Hair Brushes'': \href{https://www.amazon.com/dp/B012AE5EP4}{https://www.amazon.com/dp/B012AE5EP4}; 
(2) an eye shadow product assigned under ``Travel Cases'': \href{https://www.amazon.com/dp/B07BBM5B33}{https://www.amazon.com/dp/B07BBM5B33}. Screenshots of these product profiles are taken in 12/2019 and available at the Appendix.}

To the best of our knowledge, TXtract is the first deep neural network that has been applied to attribute value extraction for hierarchical taxonomies with thousands of product categories. In particular, we make three contributions.
\begin{enumerate}
    \item We develop TXtract, a taxonomy-aware deep neural network for attribute value extraction from product profiles for multiple product categories. In TXtract, we capture the \emph{hierarchical} relations between categories into \emph{category embeddings}, which in turn we use as context to generate category-specific token embeddings via conditional self-attention.
    \item We improve attribute value extraction through multi-task learning: TXtract jointly extracts attribute values and predicts the product's categories by sharing representations across tasks.
    \item We evaluate TXtract on a taxonomy of 4,000 product categories and show that it substantially outperforms state-of-the-art models by up to 10\% in F1 and 15\% in coverage across \emph{all} product categories.
\end{enumerate}

Although this work focuses on e-Commerce, our approach to leverage taxonomies can be applied to broader domains such as finance, education, and biomedical/clinical research. We leave experiments on these domains for future work. 

The rest of this paper is organized as follows. 
Section~\ref{s:related-work} discusses related work. 
Section~\ref{s:background} presents background and formally defines the problem. 
Section~\ref{s:model} presents our solution and Section~\ref{s:experiments} describes experimental results. Finally, Section~\ref{s:conclusions} concludes and suggests future work. 

\section{Related Work}
\label{s:related-work}
Here, we discuss related work on attribute value extraction (Section~\ref{ss:related-work-attribute-value-extraction}), and multi-task learning/meta-learning (Section~\ref{ss:related-work-multi-task-learning}).

\subsection{Attribute Value Extraction from Product Profiles}
\label{ss:related-work-attribute-value-extraction}
Attribute value extraction was originally addressed with rule-based techniques~\cite{nadeau2007survey,vandic2012faceted,gopalakrishnan2012matching} followed by supervised learning techniques~\cite{ghani2006text,putthividhya2011bootstrapped,ling2012fine,petrovski2017extracting,sheth2017}.
Most recent techniques consider open attribute value extraction: emerging attribute values can be extracted by sequence tagging, similar to named entity recognition (NER)~\cite{putthividhya2011bootstrapped,chiu2016named,lample2016neural,yadav2018survey}.
State-of-the-art methods employ deep learning for sequence tagging~\cite{zheng2018opentag,xu2019scaling,rezk2019accurate}.
However, all previous methods can be adapted to a small number of categories and require many labeled datapoints per category.\footnote{\citet{zheng2018opentag} considered 3 categories: ``Dog Dood,'' ``Cameras,'' and ``Detergent.''~\citet{xu2019scaling} consider 1 category: ``Sports \& Entertainment.''~\citet{rezk2019accurate} considered 21 categories and trained a separate model for each category.}
Even the Active Learning method of~\citet{zheng2018opentag} requires humans to annotate at least hundreds of carefully selected examples per category.
Our work differs from previous approaches as we consider thousands of product categories organized in a hierarchical taxonomy.

\subsection{Multi-Task/Meta- Learning}
\label{ss:related-work-multi-task-learning}
Our framework is related to multi-task learning~\cite{caruana1997multitask} as we train a single model simultaneously on all categories (tasks). 
Traditional approaches consider a small number of different tasks, ranging from 2 to 20 and employ hard parameter sharing~\cite{alonso2017multitask,yang2017transfer,ruder2019neural}: the first layers of neural networks are shared across all tasks, while the separate layers (or ``heads'') are used for each individual task. 
In our setting with thousands of different categories (tasks), our approach is efficient as we use a \emph{single} (instead of thousands) head and effective as we distinguish between categories through low-dimensional category embeddings.  

Our work is also related to meta-learning approaches based on task embeddings~\cite{finn2017model,achille2019task2vec,lan2019meta}: the target tasks are represented in a low-dimensional space that captures task similarities. 
However, we generate category embeddings that reflect the \emph{already available, hierarchical} structure of product categories in the taxonomy provided by experts.

\section{Background and Problem Definition}
\label{s:background}
We now provide background on open attribute value extraction (Section~\ref{ss:open-attr-val-extr}) and define our problem of focus (Section~\ref{ss:problem-definition}). 

\subsection{Open Attribute Value Extraction}
\label{ss:open-attr-val-extr}
Most recent approaches for attribute value extraction rely on the open-world assumption to discover attribute values that have never been seen during training~\cite{zheng2018opentag}.
State-of-the-art approaches address extraction with deep sequence tagging models~\cite{zheng2018opentag,xu2019scaling,rezk2019accurate}: each token of the input sequence $x=(x_1,\dots,x_T)$ is assigned a separate tag from \{B, I, O, E\}, where 
``B,'' ``I,'' ``O,'' and ``E'' represent the beginning, inside, outside, and end of an attribute, respectively. 
(Not extracting any values corresponds to a sequence of ``O''-only tags.)
Table~\ref{tab:sequence-tagging-example} shows an input/output example of \emph{flavor} value extraction from (part of) a product title.
Given this output tag sequence, ``black cherry cheesecake'' is extracted as a \emph{flavor} for the ice cream product.  

\begin{table}[t]
    \centering
    \resizebox{\columnwidth}{!}{
    \begin{tabular}{|c|c|c|c|c|c|c|c|c|}
    \hline
       \textbf{Input}  & Ben & \& & Jerry's & black & cherry & cheesecake & ice & cream \\\hline
      \textbf{Output}  & O & O & O & B & I & E & O & O \\\hline
    \end{tabular}}
    \caption{Example of input/output tag sequences for the ``flavor'' attribute of an ice cream product.}
    \label{tab:sequence-tagging-example}
\end{table}

\subsection{Problem Definition}
\label{ss:problem-definition}
We represent the product taxonomy as a tree $C$, where the root node is named ``Product'' and each taxonomy node corresponds to a distinct product category: $c \in C$.
A directed edge between two nodes represents the category-to-subcategory relationship. A product is assigned to a category node in $C$. In practice, there are often thousands of nodes in a taxonomy tree and the category assignment of a product may be incorrect. We now formally define our problem as follows.

\paragraph{DEFINITION:} Consider a product from a category $c$ and the sequence of tokens $x=(x_1,\dots,x_T)$ from its profile, where $T$ is the sequence length. Let $a$ be a target attribute for extraction. Attribute extraction identifies sub-sequences of tokens from $x$, each sub-sequence representing a value for $a$.

For instance, given (1) a product title  $x=$``Ben \& Jerry's Strawberry Cheesecake Ice Cream 16 oz,'' (2) a product category $c=$ ``Ice Cream,'' and (3) a target attribute $\alpha=$ \emph{flavor}, we would like to extract ``Strawberry Cheesecake'' as a \emph{flavor} for this product. 
Note that \emph{we may not see all valid attribute values during training}. 

\begin{figure*}[t]
\centering
\includegraphics[scale=0.25]{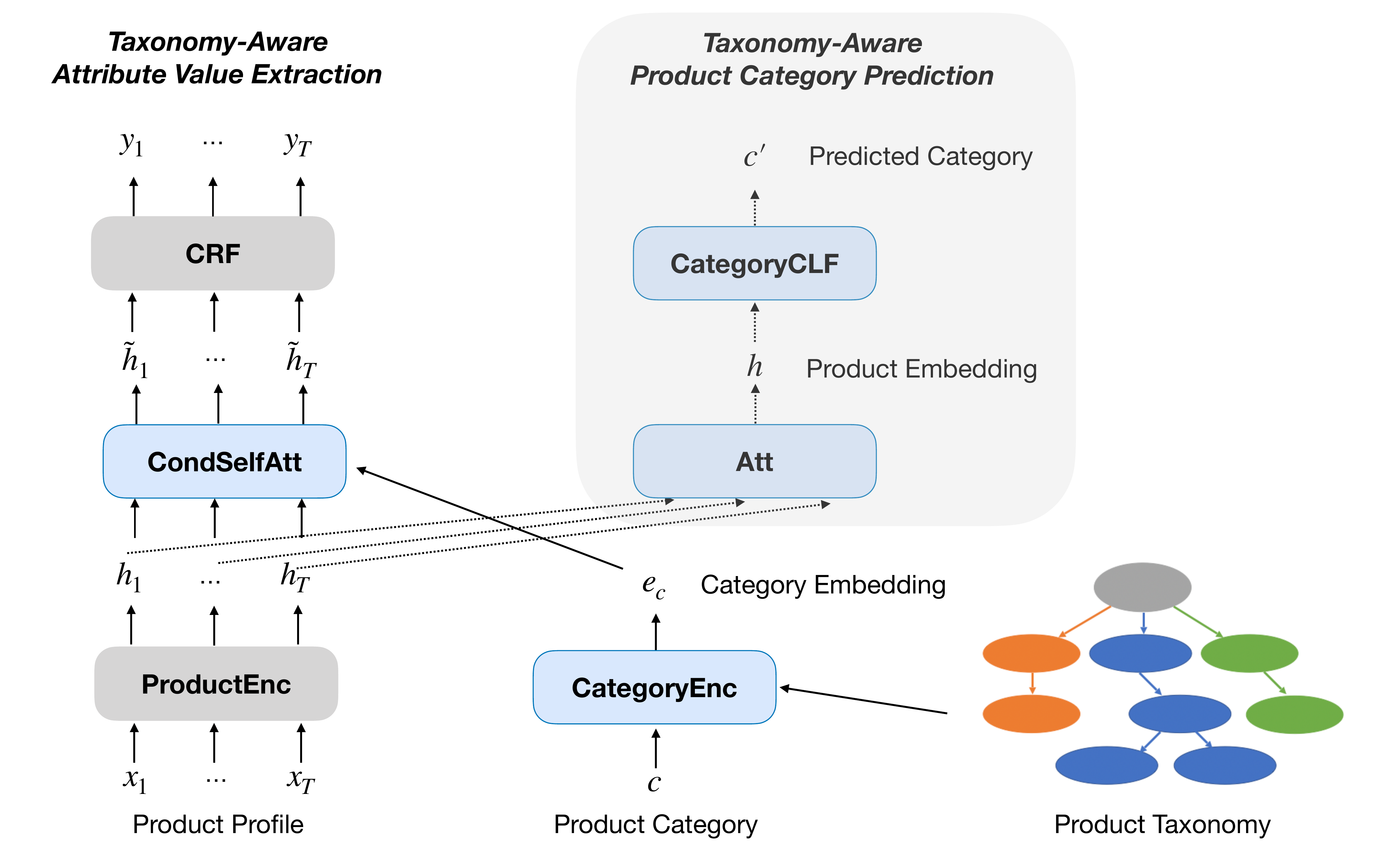}

	\caption{TXtract architecture: tokens $(x_1,\dots,x_T)$ are classified to BIOE attribute tags $(y_1, \dots, y_T)$ by conditioning to the product's category embedding $e_c$. TXtract is jointly trained to extract attribute values and assign a product to taxonomy nodes.}
\label{fig:model_architecture}
\end{figure*}  
\section{TXtract Model: Taxonomy-Aware Knowledge Extraction}
\label{s:model}
In this paper, we address open attribute value extraction using a taxonomy-aware deep sequence tagging model, TXtract. Figure~\ref{fig:model_architecture} shows the model architecture, which contains two key components: attribute value extraction and product category prediction, accounting for the two tasks in multi-task training. Both components are taxonomy aware, as we describe next in detail.

%
%

\subsection{Taxonomy-Aware Attribute Value Extraction}
\label{s:model-taxonomy-aware-extraction}
TXtract leverages the product taxonomy for attribute value extraction. The underlying intuition is that knowing the product category may help infer attribute applicability and associate the product with a certain range of valid attribute values. 
Our model uses the category embedding in conditional self-attention to guide the extraction of category-specific attribute values.

\subsubsection{Product Encoder}
\label{s:model-product-encoder}
The product encoder (``ProductEnc'') represents the text tokens of the product profile $(x_1, \dots, x_T)$ as low-dimensional, real-valued vectors:
\begin{equation}
h_1, \dots h_T = \operatorname{ProductEnc(x_1,\dots, x_T)} \in \mathbb{R}^d.    
\end{equation}
To effectively capture long-range dependencies between the input tokens, we use word embeddings followed by bidirectional LSTMs (BiLSTMs), similar to previous state-of-the-art approaches~\cite{zheng2018opentag,xu2019scaling}. 

\subsubsection{Category Encoder}
\label{ss:generating-taxonomy-embeddings}
Our category encoder (``CategoryEnc'') encodes the hierarchical structure of product categories such that TXtract understands expert-defined relations across categories, such as ``Lager'' is a sub-category of ``Beer''.
In particular, we embed each product category $c$ (taxonomy node) into a low-dimensional latent space:
\begin{equation}
    e_c = \operatorname{CategoryEnc}(c) \in \mathbb{R}^m.
\end{equation}
To capture the hierarchical structure of the product taxonomy, we embed product categories into the $m$-dimensional Poincaré ball~\cite{nickel2017poincare}, because its underlying geometry has been shown to be appropriate for capturing both similarity and hierarchy. 
%
%
%

%

\subsubsection{Category Conditional Self-Attention}
\label{ss:conditional-self-attention}
The key component for taxonomy-aware value extraction is category conditional self-attention (``CondSelfAtt''). 
CondSelfAtt generates category-specific token embeddings ($\tilde h_i \in \mathbb{R}^d$) by conditioning on the category embedding $e_c$: 
\begin{equation}
    \tilde h_1, \dots \tilde h_T = \operatorname{CondSelfAtt}((h_1, \dots, h_T), e_c).
\end{equation}

To leverage the mutual interaction between all pairs of token embeddings $h_t$, $h_{t'}$ and the category embedding $e_{c}$ we use self-attention and compute pairwise sigmoid attention weights: 
\begin{equation}
    \alpha_{t,t'} = \sigma(w_\alpha^T g_{t,t'} + b_\alpha), \qquad t,t' = 1..T.
\end{equation}
We compute scores $g_{t,t'}$ using both the token embeddings $h_t$, $h_{t'}$ and the category embedding $e_c$: 
\begin{equation}
    g_{t,t'} = \tanh (W_1 h_t + W_2 h_{t'} + W_3 e_{c} + b_g),
\end{equation}
where $W_1 \in \mathbb{R}^{p \times d}$, $W_2 \in \mathbb{R}^{p \times d}$, $W_3 \in \mathbb{R}^{p \times m}$, $w_\alpha \in \mathbb{R}^{p}$ are trainable attention matrices and $b_g \in \mathbb{R}^p$, $b_\alpha \in \mathbb{R}$, are trainable biases. 
The $T \times T$ attention matrix $A={a_{t,t'}}$ stores the pairwise attention weights. 
The contextualized token embeddings are computed as: 
\begin{equation}
    \tilde h_t = \sum_{t'=1}^T \alpha_{t,t'} \cdot h_{t'}.
\end{equation}

\subsubsection{CRF Layer}
\label{ss:model-crf-layer}
We feed the contextualized token representations $\tilde h=(\tilde h_1,\dots, \tilde h_T)$ to
CRFs to get the sequence of BIOE tags with the highest probability: 
\begin{equation}
    y_1,\dots,y_T = \operatorname{CRF}(\tilde h_1,\dots,\tilde h_t).
\end{equation}
We then extract attribute values as valid sub-sequences of the input tokens ($x_1,\dots,x_T$) with B/I/E tags (see Section~\ref{ss:open-attr-val-extr}). 

\subsubsection{Training for Attribute Value Extraction}
Our training objective for attribute value extraction is to minimize the negative conditional log-likelihood of the model parameters on $N$ training products $x_i$ with ground truth labels $\hat y_{i1} \dots, \hat y_{iT}$:
\begin{equation}
    \label{eq:attribute-value-extraction-loss}
    L_a = - \sum_{i=1}^N\log Pr(\hat y_{i1},\dots,\hat y_{iT} \mid x_i, c_i)
\end{equation}
We train our model on all categories in parallel, thus leveraging for a given category products from related categories. 
To generate training sequence labels from the corresponding attribute values, we use the distant supervision framework of~\citet{mintz2009distant}, similar to~\citet{xu2019scaling}, by generating tagging labels according to existing (sparse) values in the Catalog.

\subsection{Taxonomy-Aware Product Category Prediction}
\label{ss:multi-task-learning}
We now describe how we train TXtract for the auxiliary task of product category prediction through multi-task learning. Our main idea is that by encouraging TXtract to predict the product categories using only the product profile, the model will learn token embeddings that are discriminative of the product categories. Thus, we introduce an inductive bias for more effective category-specific attribute value extraction. 

\subsubsection{Attention Layer}
Our attention component (``Att'') represents the product profile $(x_1,\dots,x_T)$ as a single vector $h \in \mathbb{R}^n$ computed through the weighted combination of the ProductEnc's embeddings $(h_1,\dots,h_T)$:
\begin{equation}
    h = \sum_{t=1}^T \beta_t \cdot h_t.
\end{equation}
This weighted combination allows tokens that are more informative for a product's category to get higher ``attention weights'' $\beta_1, \dots, \beta_T \in [0,1]$. 
For example, we expect $x_t=$ ``frozen'' to receive a relatively high $\beta_t$ for the classification of a product to the ``Ice Cream'' category. 
We compute the attention weights as:
\begin{equation}
    \beta_t = \operatorname{softmax}(u_c^T \tanh(W_c h_t + b_c)),
\end{equation}
where $W_c \in \mathbb{R}^{q\times d}$, $b_c \in \mathbb{R}^q$, $u_c \in \mathbb{R}^q$ are trainable attention parameters.

\subsubsection{Category Classifier}
Our category classifier (``CategoryCLF'') classifies the product embedding $h$ to the taxonomy nodes. 
In particular, we use a sigmoid classification layer to predict the probabilities of the taxonomy nodes: 
\begin{equation}
   p_1,\dots,p_{|C|} = \operatorname{sigmoid}(W_d h + b_d),
\end{equation}
where $W_d \in \mathbb{R}^{|C|\times d}$ and $b_d \in \mathbb{R}^{|C|}$ are trainable parameters.
We compute sigmoid (instead of softmax) node probabilities because we treat category prediction as \emph{multi-label} classification, as we describe next.

\subsubsection{Training for Category Prediction}
\label{ss:category-prediction-training}
Training for ``flat'' classification of products to thousands of categories is not effective because the model is fully penalized if it does not predict the exact true category $\hat c$ while at the same time ignores parent-children category relations.
Here, we conduct ``hierarchical'' classification by incorporating the hierarchical structure of the product taxonomy into a \emph{taxonomy-aware} loss function.

The insight behind our loss function is that a product assigned under $\hat c$ could also be assigned under any of the ancestors of $\hat c$.
Thus, we consider hierarchical multi-label classification and encourage TXtract to assign a product to all nodes in the path from $\hat c$ to the root, denoted by ($\hat c_K, \hat c_{K-1}, \dots, \hat c_1$), where $K$ is the level of the node $\hat c$ in the taxonomy tree. The model is thus encouraged to learn the hierarchical taxonomy relations and will be penalized less if it predicts high probabilities for ancestor nodes (e.g., "Beer" instead of ``Lager'' in Figure~\ref{fig:amazon_product_example}).

Our minimization objective is the \emph{weighted} version of the binary cross-entropy (instead of \emph{unweighted} categorical cross-entropy) loss:\footnote{For simplicitly in notation, we define Eq~\ref{eq:hierarchical-loss-function} for a single product. Defining for all training products is straightforward.} 
\begin{equation}
\label{eq:hierarchical-loss-function}
    L_b = \sum_{c \in C} w_{c} (y_{c} \cdot \log p_{c} + (1-y_{c})\cdot \log (1-p_{c})), 
\end{equation}
For the nodes in the path from $\hat c$ to the root ($\hat c_K, \hat c_{K-1}, \dots, \hat c_1$), we define positive labels $y_c=1$ and weights $w_c$ that are exponentially decreasing ($w^0$, $w^1, \dots, w^{K-1}$), where $0 < w \leq 1$ is a tunable hyper-parameter.
The remaining nodes in $C$ receive negative labels $y_c=0$ and fixed weight $w_c=w^{K-1}$.

\subsection{Multi-task Training}
We jointly train TXtract for attribute value extraction and product category prediction by combining the loss functions of Eq.~\eqref{eq:attribute-value-extraction-loss} and Eq.~\eqref{eq:hierarchical-loss-function}:
\begin{equation}
    L = \gamma \cdot L_a + (1-\gamma) \cdot L_b,
\end{equation}
where $\gamma \in [0,1]$ is a tunable hyper-parameter.
Here, we employ multi-task learning, and share ProductEnc across both tasks. %

\section{Experiments}
\label{s:experiments}
We empirically evaluated TXtract and compared it with state-of-the-art models and strong baselines for attribute value extraction on 4000 product categories. 
TXtract leads to substantial improvement across all categories, showing the advantages of leveraging the product taxonomy. 

\begin{table*}[t]
    \centering
    \resizebox{2.1\columnwidth}{!}{
    \begin{tabular}{|l|l||l|l||l|l|l||l|l|l|}
    \hline
       \textbf{Attr.}  &  \textbf{Model} & \textbf{Vocab} &\textbf{Cov} & \textbf{Micro F1} &  \textbf{Micro  Prec} &  \textbf{Micro Rec} & \textbf{Macro F1} & \textbf{Macro Prec} &  \textbf{Macro Rec}\\\hline
       \multirow{ 2}{*}{\emph{Flavor}}  & OpenTag  & 6,756& 73.2  &  57.5 & 70.3	& 49.6 & 54.6 & 68.0 & 47.3\\
         & TXtract  & \textbf{13,093}&\textbf{83.9} \small{$\uparrow$14.6\%} &  \textbf{63.3} \small{$\uparrow$10.1\%} & \textbf{70.9}  \small{$\uparrow$0.9\%}	&\textbf{57.8}  \small{$\uparrow$16.5\%} & \textbf{59.3} \small{$\uparrow$8.6\%} & \textbf{68.4} \small{$\uparrow$0.6\%} &	\textbf{53.8} \small{$\uparrow$13.7\%}\\\hline
       \multirow{ 2}{*}{\emph{Scent}}  & OpenTag  & 10,525 & 75.8  &  70.6 &\textbf{87.6} & 60.2 & 59.3 & \textbf{79.7} &	50.8\\
         & TXtract  & \textbf{13,525} &\textbf{83.2} \small{$\uparrow$9.8\%} &\textbf{73.7} \small{$\uparrow$4.4\%}  & 86.1 \small{$\downarrow$1.7\%}&	\textbf{65.7} \small{$\uparrow$9.1\%}& \textbf{59.9} \small{$\uparrow$10.1\%}  & 78.3 \small{$\downarrow$1.8\%}& \textbf{52.1} \small{$\uparrow$2.6\%}\\\hline
       \multirow{ 2}{*}{\emph{Brand}}  & OpenTag  & 48,943 & 73.1   & 63.4  &81.6& 51.9 & 51.7  & 75.1 &41.5\\
         & TXtract  & \textbf{64,704} & \textbf{82.9} \small{$\uparrow$13.4\%} &  \textbf{67.5} \small{$\uparrow$6.5\%}  & \textbf{82.7} \small{$\uparrow$1.3\%} & \textbf{56.5} \small{$\uparrow$8.1\%} & \textbf{55.3} \small{$\uparrow$7.0\%}  &  \textbf{75.2} \small{$\uparrow$0.1\%}& \textbf{46.8} \small{$\uparrow$12.8\%}\\\hline
       \multirow{ 2}{*}{\emph{Ingred.}}  & OpenTag  & 9,910& 70.0& 35.7 & 46.6 &	29.1 & 20.9  & 34.6 & 16.7 \\
        & TXtract  &  \textbf{18,980} & \textbf{76.4} \small{$\uparrow$9.1\%}& \textbf{37.1} \small{$\uparrow$3.9\%} & \textbf{48.3} \small{$\uparrow$3.6\%} &	\textbf{30.1} \small{$\uparrow$3.3\%}& \textbf{24.2} \small{$\uparrow$15.8\%}  & \textbf{37.4} \small{$\uparrow$8.1\%}&	\textbf{19.8} \small{$\uparrow$18.6\%}\\\hline\hline
         \multicolumn{3}{|c|}{Average relative increase} & $\uparrow$11.7\% & $\uparrow$6.2\% & $\uparrow$1.0\% & $\uparrow$9.3\% & $\uparrow$10.4\% & $\uparrow$6.8\% & $\uparrow$11.9\%\\\hline
    \end{tabular}}
    \caption{Extraction results for \emph{flavor}, \emph{scent}, \emph{brand}, and \emph{ingredients} across 4,000 categories. Across all attributes, TXtract improves OpenTag by 11.7\% in coverage, 6.2\% in micro-average F1, and 10.4\% in macro-average F1.}
    \label{tab:results-all-attributes}
\end{table*}

\subsection{Experimental Settings}
\paragraph{Dataset:} 
We trained and evaluated TXtract on products from public web pages of Amazon.com.
We randomly selected 2 million products from 4000 categories under 4 general domains (sub-trees) in the product taxonomy: Grocery, Baby product, Beauty product, and Health product.

\paragraph{Experimental Setup:} We split our dataset into training (60\%), validation (20\%), and test (20\%) sets. 
We experimented with extraction of \emph{flavor}, \emph{scent}, and \emph{brand} values from product titles, and with \emph{ingredient} values from product titles and descriptions. %
For each attribute, we trained TXtract on the training set and evaluated the performance on the held-out test set. 

\paragraph{Evaluation Metrics:} For a robust evaluation of attribute value extraction, we report several metrics.
For a test product, we consider as true positive the case where the extracted values match at least one of the ground truth values (as some of the ground truth values may not exist in the text) and do not contain any wrong values.\footnote{For example, if the ground-truth is [$v_1$] but the system extracts [$v_1$, $v_2$, $v_3$], the extraction is considered as incorrect.}
We compute \emph{Precision} (Prec) as the number of ``matched'' products divided by the number of products for which the model extracts at least one attribute value; \emph{Recall} (Rec) as the number of ``matched'' products divided by the number of products associated with attribute values; and \emph{F1} score as the harmony mean of Prec and Rec.
To get a global picture of the model's performance, we consider micro-average scores (Mi*), which first aggregates products across categories and computes Prec/Rec/F1 globally.
To evaluate per-category performance we consider macro-average scores (Ma*), which first computes Prec/Rec/F1 for each category and then aggregates per-category scores.
To evaluate the capability of our model to discover (potentially new) attribute values, we also report the \emph{Value vocabulary} (Vocab) as the total number of unique attribute values extracted from the test set (higher number is often better); and \emph{Coverage} (Cov), as the number of products for which the model extracted at least one attribute value, divided by the total number of products.

For product category (multi-label) classification we reported the area under Precision-Recall curve (AUPR), Precision, Recall, and F1 score.

\paragraph{Model Configuration:} We implemented our model in Tensorflow~\cite{abadi2016tensorflow} and Keras.\footnote{https://keras.io/} 
For a fair comparison, we consider the same configuration as OpenTag for the ProductEnc (BiLSTM)\footnote{We expect to see further performance improvement by considering pre-trained language models~\cite{radford2018improving, devlin2019bert} for ProductEnc, which we leave for future work.} and CRF components. For model configuration details see the appendix.

\paragraph{Model Comparison: }
\label{ss:model-comparison} We compared our model with state-of-the-art models in the literature and introduced additional strong baselines:

\begin{enumerate}
    \item ``OpenTag'': the model of~\citet{zheng2018opentag}. It is a special case of our system that consists of the ProductEnc and CRF components without leveraging the taxonomy. 
    \item ``Title+*'': a class of models for conditional attribute value extraction, where the taxonomy is introduced by artificially appending extra tokens $x_1',\dots, x_{T'}'$ and a special separator token (<$\operatorname{SEP}$>) to the beginning of a product's text, similar to~\citet{johnson2017google}:\\
   \centerline{$x' = (x_1',\dots, x_{T'}'$, <$\operatorname{SEP}$>, $x_1, \dots, x_T)$}
    Tokens $x_1',\dots, x_{T'}'$ contain category information such as unique category id (``Title+id''), category name (``Title+name''), or the names of all categories in the path from the root to the category node, separated by an extra token <$\operatorname{SEP2}$> (``Title+path'').
    \item ``Concat-*'': a class of models for taxonomy-aware attribute value extraction that concatenate the category embedding to the word embedding (-wemb) or hidden BiLSTM embedding layer (-LSTM) instead of using conditional self-attention. 
    We evaluate Euclidean embeddings (``Concat-*-Euclidean'') and Poincaré embeddings (``Concat-*-Poincaré''). %
    \item ``Gate'': a model that leverages category embeddings $e_c$ in a gating layer~\cite{cho2014learning, ma2019hierarchical}: 
    $ \tilde h_t = h_t \otimes \sigma(W_4 h_t + W_5 e_{c}),$
    where $W_4 \in \mathbb{R}^{p \times d}$, $W_5 \in \mathbb{R}^{p \times m}$ are trainable matrices, and $\otimes$ denotes element-wise multiplication.  
    Our conditional self-attention is different as it leverages pairwise instead of single-token interactions with category embeddings.
    \item ``CondSelfAtt'': the model with our conditional self-attention mechanism (Section~\ref{ss:conditional-self-attention}). 
    CondSelfAtt extracts attribute values but does not predict the product category.
    \item ``MT-*'': a multi-task learning model that jointly performs (\emph{not} taxonomy-aware) attribute value extraction and category prediction. ``MT-flat'' assumes ``flat'' categories, whereas ``MT-hier'' considers the hierarchical structure of the taxonomy (Section~\ref{ss:category-prediction-training}).
    \item ``TXtract'': our model that jointly performs \emph{taxonomy-aware} attribute value extraction (same as CondSelfAtt) and \emph{hierarchical} category prediction (same as MT-hier). 
\end{enumerate}

Here, we do not report previous models (e.g., BiLSTM-CRF) for sequence tagging~\cite{huang2015bidirectional,kozareva2016recognizing,lample2016neural}, as OpenTag has been shown to outperform these models in~\citet{zheng2018opentag}. 
Moreover, when considering attributes separately, the model of~\citet{xu2019scaling} is the same as OpenTag, but with a different ProductEnc component; since we use the same ProductEnc for all alternatives, we expect/observe the same trend and do not report its performance.

\subsection{Experimental Results}
\label{ss:experimental-results}
Table~\ref{tab:results-all-attributes} reports the results across all categories.
For detailed results see Figure~\ref{tab:detailed-results-per-category} in Appendix.
Over all categories, our taxonomy-aware TXtract substantially improves over the state-of-the-art OpenTag by up to 10.1\% in Micro F1, 14.6\% in coverage, and 93.8\% in vocabulary (for \emph{flavor}).

Table~\ref{tab:coarse-vs-fine-grained-results} shows results for the four domains of our taxonomy under different training granularities: training on all domains versus training only on the target domain.  
Regardless of the configuration, TXtract substantially outperforms OpenTag, showing the general advantages of our approach. Interestingly, although training a single model on all of the four domains obtains lower F1 for {\em Flavor}, it obtains better results for {\em Scent}: training fewer models does not necessarily lead to lower quality and may actually improve extraction by learning from neighboring taxonomy trees.

\begin{table}[t]
    \centering
    \resizebox{\columnwidth}{!}{
    \begin{tabular}{|c|c|c||c|}
    \hline
         \multicolumn{2}{|c|}{\textbf{Domain}} & &  \textbf{OpenTag/TXtract}\\

       \textbf{Train} & \textbf{Test} &\textbf{Attr.} &   \textbf{Micro F1}\\\hline
        all & Grocery & \multirow{ 2}{*}{\emph{Flavor}}  & 60.3 / \textbf{64.9} \small{$\uparrow$7.6\%} \\
         Grocery & Grocery & & 65.4 / \textbf{70.5} \small{$\uparrow$7.8\%}\\\hline
        all & Baby & \multirow{ 2}{*}{\emph{Flavor}}  & 54.4 / \textbf{63.0} \small{$\uparrow$15.8\%}\\
        Baby & Baby & & 69.2 / \textbf{71.8} \small{$\uparrow$3.8\%} \\\hline
     all & Beauty & \multirow{ 2}{*}{\emph{Scent}} & 76.9 / \textbf{79.5} \small{$\uparrow$3.4\%}\\
     Beauty & Beauty & & 76.9 / \textbf{79.0} \small{$\uparrow$2.7\%}\\\hline 
     all & Health & \multirow{ 2}{*}{\emph{Scent}} & 63.0 / \textbf{69.1} \small{$\uparrow$9.7\%}\\
     Health & Health & & 60.9 / \textbf{63.5} \small{$\uparrow$4.3\%}\\\hline 
    \end{tabular}}
    \caption{Evaluation results for each domain under training configurations of different granularity. TXtract outperforms OpenTag under all configurations.}
    \label{tab:coarse-vs-fine-grained-results}
\end{table}

\subsection{Ablation Study}
Table~\ref{tab:flavor-results-detailed} reports the performance of several alternative approaches for \emph{flavor} value extraction across all categories. 
OpenTag does not leverage the product taxonomy, so it is outperformed by most approaches that we consider in this work. 

\paragraph{Implicit vs. explicit conditioning on categories.}
``Title+*'' baselines fail to leverage the taxonomy, thus leading to lower F1 score than OpenTag: implicitly leveraging categories as artificial tokens appended to the title is not effective in our setting.

Representing the taxonomy with category embeddings leads to significant improvement over OpenTag and ``Title+*'' baselines: 
even simpler approaches such as ``Concat-*-Euclidean'' outperform OpenTag across all metrics.
However, ``Concat-*'' and ``Gate-*'' do not leverage category embeddings as effectively as ``CondSelfAtt'': conditioning on the category embedding for the computation of the pair-wise attention weights in the self-attention layer appears to be the most effective approach for leveraging the product taxonomy.

\begin{table}[t]
\resizebox{\columnwidth}{!}{
\begin{tabular}{|l|l|l||l|}
\hline
\textbf{Model} & \textbf{TX}      & \textbf{MT} & \textbf{Micro F1} \\\hline
OpenTag          & -&-              &  57.5 \\\hline
Title+id    & \checkmark &  -   & 55.7 \small{$\downarrow$3.1\%} \\
Title+name   & \checkmark & - &  56.9 \small{$\downarrow$1.0\%}   \\
Title+path    & \checkmark &  -   & 54.3 \small{$\downarrow$5.6\%} \\
Concat-wemb-Euclidean  & \checkmark &- &  60.1 \small{$\uparrow$4.5\%} \\
Concat-wemb-Poincaré   & \checkmark &  -&  60.6 \small{$\uparrow$5.4\%} \\
Concat-LSTM-Euclidean  & \checkmark &- &  60.1 \small{$\uparrow$4.5\%} \\
Concat-LSTM-Poincaré  & \checkmark &- &  60.8 \small{$\uparrow$5.7\%} \\
Gate-Poincaré     & \checkmark &  - & 60.6  \small{$\uparrow$5.4\%}\\
CondSelfAtt-Poincaré  & \checkmark & - & 61.9 \small{$\uparrow$7.7} \\\hline
MT-flat             & - &      \checkmark  & 60.9 \small{$\uparrow$5.9\%} \\
MT-hier                & - &      \checkmark &61.5 \small{$\uparrow$7.0\%}\\\hline
Concat \& MT-hier    &  \checkmark &      \checkmark & 62.3 \small{$\uparrow$8.3\%}\\
Gate \& MT-hier  &  \checkmark &      \checkmark & 61.1 \small{$\uparrow$6.3\%}\\
CondSelfAtt \& MT-hier &  \checkmark &      \checkmark & \textbf{63.3} \small{$\uparrow$10.1\%} \\ 
\hline 
\end{tabular}}
\caption{Ablation study for \emph{flavor} extraction across 4,000 categories. ``TX'' column indicates whether the taxonomy is leveraged for attribute value extraction (Section~\ref{s:model-taxonomy-aware-extraction}). ``MT'' column indicates whether multi-task learning is used (Section~\ref{ss:multi-task-learning}).}
\label{tab:flavor-results-detailed}
\end{table}

\begin{table}[t]
    \centering
    \resizebox{\columnwidth}{!}{
    \begin{tabular}{|c||c|c|c|c|}
    \hline
        \textbf{Category Prediction} & \textbf{AUPR} & \textbf{F1} & \textbf{Prec} & \textbf{Rec}  \\\hline
         Flat  & 0.61 & 53.9 &74.2	 &48.0	  \\
         Hierarchical  & \textbf{0.68} & \textbf{62.7} &\textbf{80.4} 	& \textbf{56.9} \\
    \hline
    \end{tabular}}
    \caption{Performance of product classification to the 4,000 nodes in the taxonomy using flat versus hierarchical multi-task learning.}
    \label{tab:flat-vs-hierarchical-prediction}
\end{table}
\paragraph{Multi-task Learning.}
In Table~\ref{tab:flavor-results-detailed}, both MT-flat and MT-hier, which do not condition on the product taxonomy, outperform OpenTag on attribute value extraction: by learning to predict the product category, our model implicitly learns to condition on the product category for effective attribute value extraction.  
MT-hier outperforms MT-flat: leveraging the hierarchical structure of the taxonomy is more effective than assuming flat categories. 
Table~\ref{tab:flat-vs-hierarchical-prediction} shows that category prediction is more effective when considering the hierarchical structure of the categories into our taxonomy-aware loss function than assuming flat categories.
\begin{figure*}[t]
    \centering
    \begin{subfigure}[t]{0.49\textwidth}
        \centering
        \includegraphics[height=5cm]{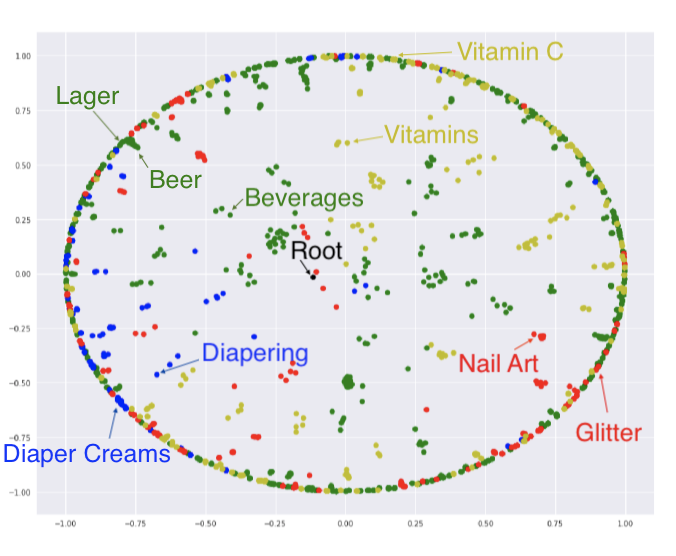}
        \caption{Taxonomy embeddings in the 2-dimensional Poincaré disk, where the distance of points grows exponentially to the radius. Leaf nodes are placed close to the boundary of the disk.}
        \label{fig:poincare_embeddings_2d}
    \end{subfigure}%
    ~ 
    \begin{subfigure}[t]{0.49\textwidth}
        \centering
        \includegraphics[height=5cm]{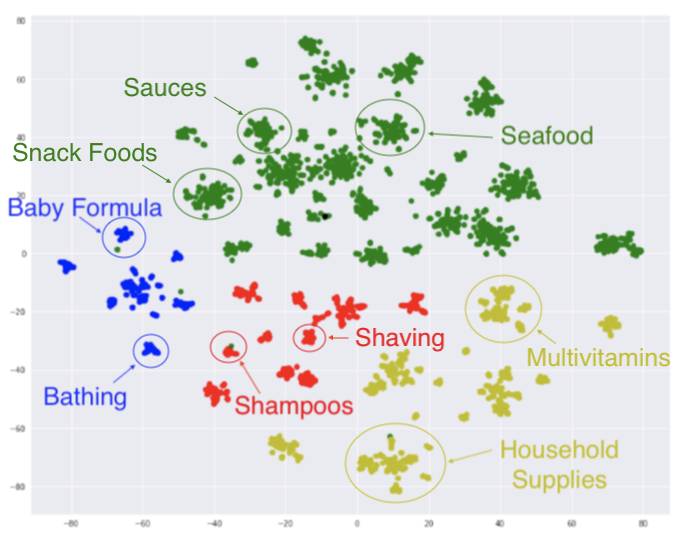}
        \caption{Taxonomy embeddings projected from the 50-dimensional Poincaré ball to the 2-dimensional Euclidean space using t-SNE. Small clusters correspond to taxonomy sub-trees.}
        \label{fig:poincare_embeddings_50d}
    \end{subfigure}
    \caption{Poincaré embeddings of taxonomy nodes (product categories). Each point is a product category. Categories are colored based on the first-level taxonomy where they belong (green: Grocery products, blue: Baby products, red: Beauty products, yellow: Health products). Related categories in the taxonomy (e.g., categories belonging to the same sub-tree) have similar embeddings.}
    \label{fig:poincare_embeddings_vis}
\end{figure*}

\begin{figure*}[t!]
    \centering
    \begin{subfigure}[t]{0.49\textwidth}
        \centering
        \includegraphics[width=0.99\columnwidth, height=27mm]{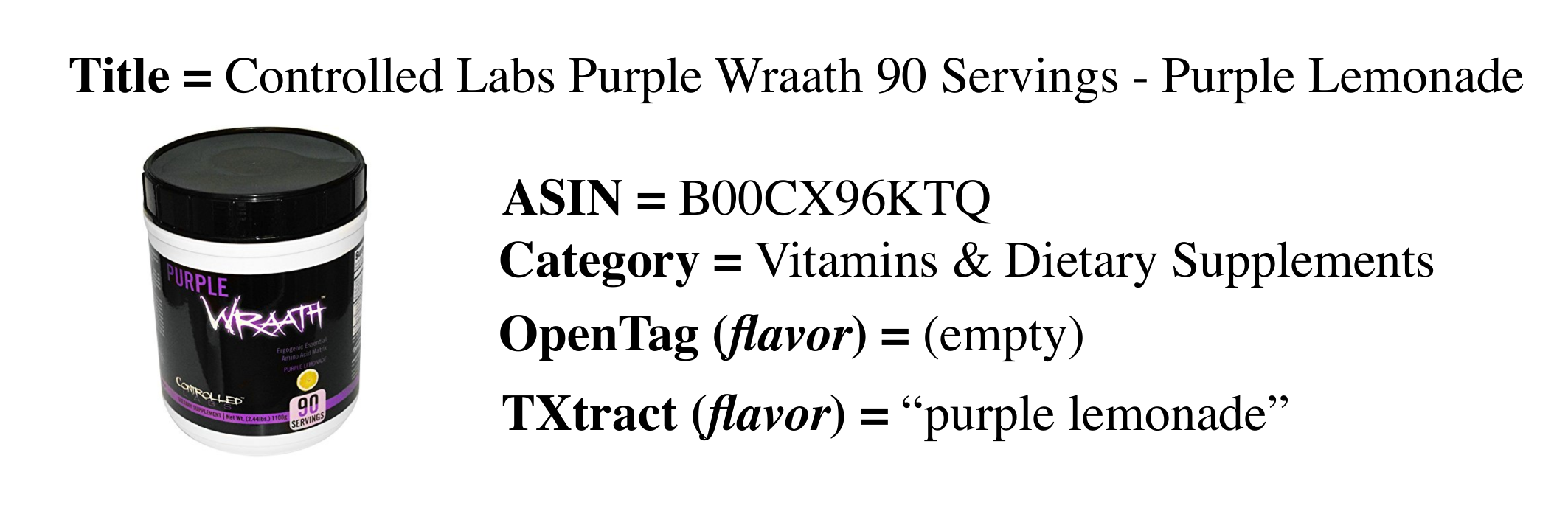}
        \caption{}
        \label{fig:example1}
    \end{subfigure}%
    ~ 
    \begin{subfigure}[t]{0.49\textwidth}
        \centering
        \includegraphics[width=0.99\columnwidth, height=27mm]{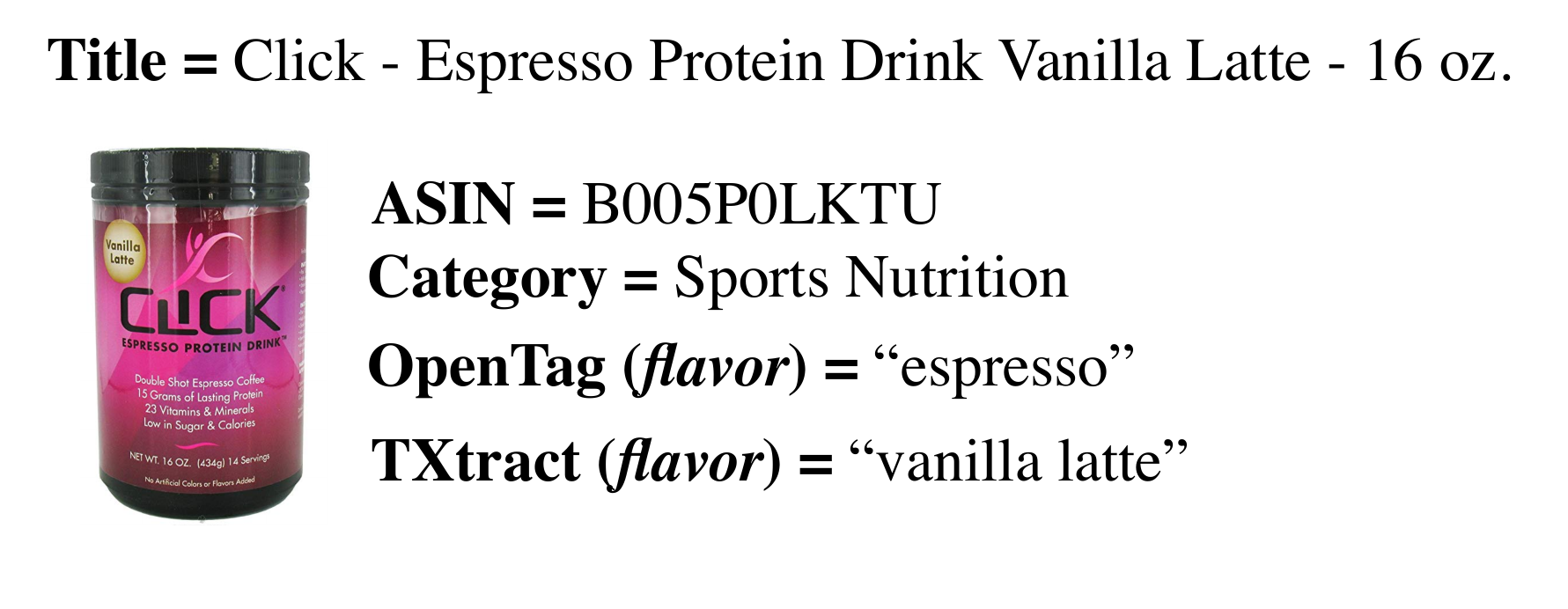}
        \caption{}
    \end{subfigure}%
    
    \begin{subfigure}[t]{0.49\textwidth}
        \centering
        \includegraphics[width=\columnwidth, height=25mm]{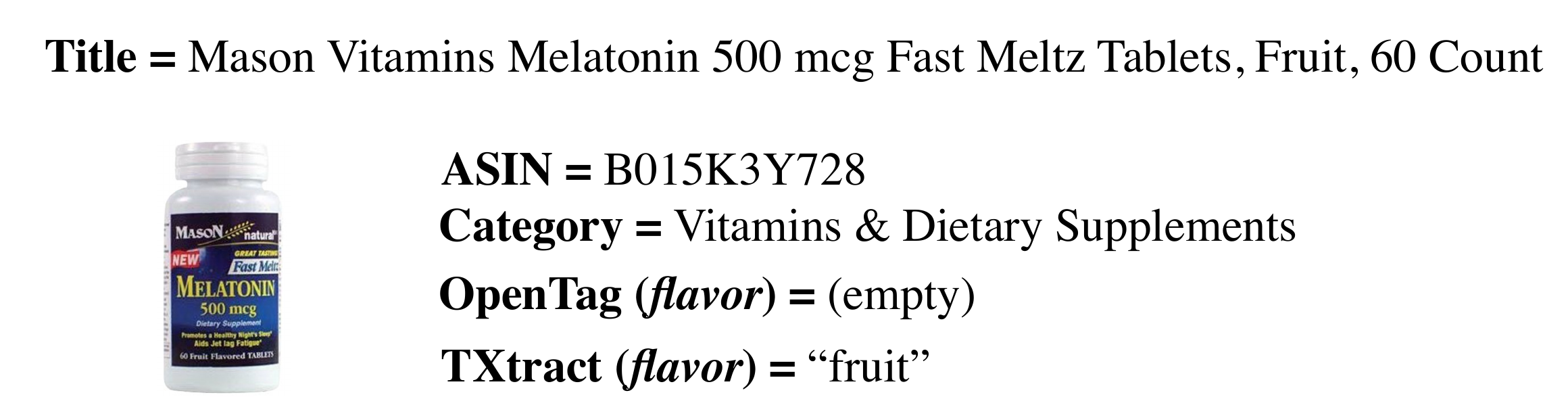}
        \caption{}
    \end{subfigure}
    ~
    \begin{subfigure}[t]{0.49\textwidth}
        \centering
        \includegraphics[width=0.99\columnwidth, height=27mm]{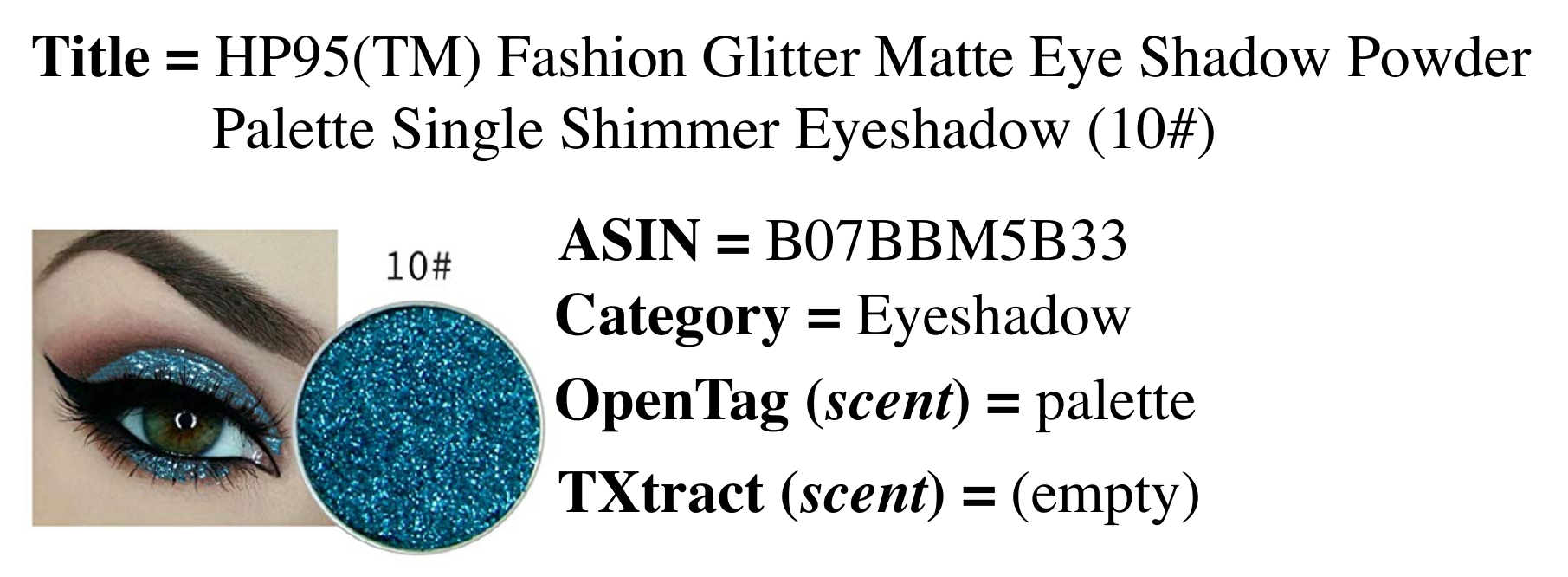}
        \caption{}
        \label{fig:example4}
    \end{subfigure}%

    \caption{Examples of extracted attribute values from OpenTag and TXtract.}
    \label{fig:examples}
\end{figure*}

\subsection{Visualization of Poincaré Embeddings}
Poincaré embeddings effectively capture the hierarchical structure of the product taxonomy: Figure~\ref{fig:poincare_embeddings_2d} plots the embeddings of product categories in the 2-dimensional Poincaré disk.\footnote{We train 2-dimensional Poincaré embeddings only for visualization. In our experiments we use $d=50$ dimensions.} 
Figure~\ref{fig:poincare_embeddings_50d} plots the embeddings trained in the 50-dimensional Poincaré ball and projected to the 2-dimensional Euclidean space through t-SNE~\cite{maaten2008visualizing}.

\subsection{Examples of Extracted Attribute Values}
Figure~\ref{fig:examples} shows examples of product titles and attribute values extracted by OpenTag or TXtract. 
TXtract is able to detect category-specific values: in Figure~\ref{fig:example1}, ``Purple Lemonade'' is a valid \emph{flavor} for ``Vitamin Pills'' but not for most of other categories. OpenTag, which ignores product categories, fails to detect this value while TXtract successfully extracts it as a \emph{flavor}.
TXtract also learns attribute applicability: in Figure~\ref{fig:example4}, OpenTag erroneously extracts ``palette'' as \emph{scent} for an ``Eyeshadow'' product, while this product should not have \emph{scent};
on the other hand, TXtract, which considers category embeddings, does not extract any \emph{scent} values for this product.

\section{Conclusions and Future Work}
\label{s:conclusions}
We present a novel method for large-scale attribute value extraction for products from a taxonomy with thousands of product categories. Our proposed model, TXtract, is 
both efficient and effective: it leverages the taxonomy into a deep neural network to improve extraction quality and can extract attribute values on all categories in parallel. 
TXtract significantly outperforms state-of-the-art approaches and strong baselines under a taxonomy with thousands of product categories.
Interesting future work includes applying our techniques to different taxonomies (e.g., biomedical) and training a model for different attributes. 

\section*{Acknowledgments}
The authors would like to sincerely thank Ron Benson, Christos Faloutsos, Andrey Kan, Yan Liang, Yaqing Wang, and Tong Zhao for their insightful comments on the paper, and Gabriel Blanco, Alexandre Manduca, Saurabh Deshpande, Jay Ren, and Johanna Umana for their constructive feedback on data integration for the experiments.

\newpage 
\bibliography{myreferences}
\bibliographystyle{acl_natbib}
\newpage
\clearpage

\appendix
\section{Appendix}
For reproducibility, we provide details
on TXtract configuration (Section~\ref{appendix:txtract-config}). 
We also report detailed evaluation results (Section~\ref{appendix:extra-results}).

\subsection{TXtract Configuration}
\label{appendix:txtract-config}
We implemented our model in Tensorflow~\cite{abadi2016tensorflow} and Keras.\footnote{https://keras.io/} 
To achieve a fair comparison with OpenTag~\cite{zheng2018opentag}, and to ensure that performance improvements stem from leveraging the product taxonomy, we use exactly the same components and configuration as OpenTag for ProductEnc: 
We initialize the word embedding layer using 100-dimensional pre-trained Glove embeddings~\cite{pennington2014glove}.
We use masking to support variable-length input. 
Each of the LSTM layers has a hidden size of 100 dimensions, leading to a BiLSTM layer with $d=200$ dimensional embeddings. We set the dropout rate to 0.4. 
For CategoryEnc, we train $m=50$-dimensional Poincaré embeddings.\footnote{We use the public code in provided by~\citet{nickel2017poincare}: \href{https://github.com/facebookresearch/poincare-embeddings}{https://github.com/facebookresearch/poincare-embeddings}}
For CondSelfAtt, we use $p=50$ dimensions. 
For Att, we use $q=50$ dimensions. 
 For multi-task training, we obtain satisfactory performance with default hyper-parameters $\gamma=0.5$, $w=1$, while we leave fine-tuning for future work.
For parameter optimization, we use Adam~\cite{kingma2014adam} with a batch size of 32.
We train our model for up to 30 epochs and quit training if the validation loss does not decrease for more than 3 epochs. 

\subsection{Extra Results}
\label{appendix:extra-results}
Table~\ref{tab:detailed-results-per-category} reports extraction results (of TXtract trained on all domains) for each domain separately.
Table~\ref{tab:product_cat_classification_detailed} reports category classification results for each domain separately. 
Table~\ref{tab:flavor-results} reports several evaluation metrics for our ablation study.

\begin{table*}[t]
    \centering
    \resizebox{2\columnwidth}{!}{
    \begin{tabular}{|c|c||c|c|c|c||c|c|c|c||c|c|c|c||c|c|c|c|}
    \hline
         &   &  \multicolumn{4}{c||}{\textbf{Grocery Products}} &  \multicolumn{4}{c||}{\textbf{Baby Products}} &  \multicolumn{4}{c||}{\textbf{Beauty Products}} &  \multicolumn{4}{c|}{\textbf{Health Products}} \\\hline
       \textbf{Attr.}  &  \textbf{Model} & \textbf{Vocab} &\textbf{Cov} & \textbf{miF1} & \textbf{maF1} & \textbf{Vocab} &\textbf{Cov} & \textbf{miF1} & \textbf{maF1} & \textbf{Vocab} &\textbf{Cov} & \textbf{miF1} & \textbf{maF1} & \textbf{Vocab} &\textbf{Cov} & \textbf{miF1} & \textbf{maF1}\\\hline\hline
       \multirow{ 2}{*}{\emph{Flavor}}  & OpenTag  &  4364 & 79.6  &  60.3 & 59.0 & 264 & 53.1 & 54.4 & 45.0 &832 & 45.8 & 41.1 & 32.0 &1296 & 58.2 & 53.9 & 47.0\\
         & TXtract  & \textbf{8607}&\textbf{89.1}  &  \textbf{64.9} & \textbf{62.8} & \textbf{414}& \textbf{72.8}& \textbf{63.0} & \textbf{56.1} & \textbf{1684} & \textbf{61.3} & \textbf{46.5} & \textbf{35.6} & \textbf{2388} & \textbf{71.5} & \textbf{67.3} & \textbf{57.5}\\\hline
       \multirow{ 2}{*}{\emph{Scent}}  & OpenTag  & 446 & 75.5  &  56.8 & 48.4 & \textbf{593} & 69.7 & 35.7 & 20.3 &7007 & 78.5 & 76.9 &67.9 & 2479& 68.1& 63.0 & 47.5\\
         & TXtract  & \textbf{565} &\textbf{87.4}  &  \textbf{61.2} & \textbf{51.4} & 589& \textbf{72.1} & \textbf{38.1} & \textbf{22.0}& \textbf{9048}& \textbf{85.6}& \textbf{79.5}& \textbf{68.4}& \textbf{3322}& \textbf{79.9}& \textbf{69.1}& \textbf{48.2}\\\hline
       \multirow{ 2}{*}{\emph{Brand}}  & OpenTag  & 5150 &	68.8&	62.9     & 52.7 &11166 & 72.2 & 66.0 & 54.0 & 15394 & 77.2 & 68.8 & 54.7 & 17233 & 71.2 & 57.8& 45.9\\
         & TXtract  & \textbf{6944} & \textbf{78.9} & \textbf{67.4}  & \textbf{55.1} & \textbf{14965} & \textbf{81.0} & \textbf{72.9} & \textbf{56.2} &\textbf{19821} & \textbf{85.1} & \textbf{72.7} & \textbf{57.2} &\textbf{22974} & \textbf{82.9} & \textbf{60.5} & \textbf{52.4} \\\hline
       \multirow{ 2}{*}{\emph{Ingred.}}  & OpenTag  & 3402 & 82.5& 40.5  & 30.1& 490 & 50.7 & 27.7 &22.4 & 2767 & 65.1& \textbf{33.6} & \textbf{26.8} & 3251 & 66.7 & 34.6 & 29.9\\
        & TXtract  & \textbf{6155}& \textbf{87.3} & \textbf{43.1} & \textbf{36.5} & \textbf{835}& \textbf{59.7} & \textbf{30.5} & \textbf{24.3} & \textbf{5539} & \textbf{70.6} & 32.9 & 26.6 & \textbf{6451} & \textbf{74.2} & \textbf{36.5} & \textbf{31.2}\\\hline
    \end{tabular}}
    \caption{Extraction results for \emph{flavor}, \emph{scent}, \emph{brand}, and \emph{ingredients} for each of our 4 domains (sub-trees).}
    \label{tab:detailed-results-per-category}
\end{table*}

\begin{table*}[t]
    \centering
    \resizebox{2\columnwidth}{!}{
    \begin{tabular}{|c||c|c|c|c||c|c|c|c||c|c|c|c||c|c|c|c|}
    \hline
           &  \multicolumn{4}{c||}{\textbf{Grocery Products}} &  \multicolumn{4}{c||}{\textbf{Baby Products}} &  \multicolumn{4}{c||}{\textbf{Beauty Products}} &  \multicolumn{4}{c|}{\textbf{Health Products}} \\\hline
        \textbf{MT type} & \textbf{AUPR} &\textbf{F1} & \textbf{Prec} & \textbf{Rec} & \textbf{AUPR} &\textbf{F1} & \textbf{Prec} & \textbf{Rec} & \textbf{AUPR} &\textbf{F1} & \textbf{Prec} & \textbf{Rec} & \textbf{AUPR} &\textbf{F1} & \textbf{Prec} & \textbf{Rec}\\\hline\hline
       flat  & 45.9  & 21.4  & 63.3  & 13.7 & 65.9 & 23.7 & 68.4  & 17.4  & 63.7 & 62.4 & 78.8 & 56.5  & 49.8 & 38.8 & 60.7  & 32.7 \\\hline
      hierarchical  & \textbf{47.3}  & \textbf{29.7}  & \textbf{68.4}  & \textbf{19.9} & \textbf{68.5} & \textbf{29.4} & \textbf{72.6}  & \textbf{22.9}  & \textbf{72.1} & \textbf{71.5} & \textbf{83.1} & \textbf{66.4} & \textbf{56.3} & \textbf{47.7} & \textbf{74.6} & \textbf{39.8} \\\hline

    \end{tabular}}
    \caption{Product category classification results}
    \label{tab:product_cat_classification_detailed}
\end{table*}

\begin{table*}[t]
\resizebox{2\columnwidth}{!}{
\begin{tabular}{|l|c|c||cc|ccc|ccc|}
\hline
  & & & & & \multicolumn{3}{c|}{\textbf{Micro-average}} &  \multicolumn{3}{c|}{\textbf{Macro-average}}   \\

\textbf{Model} & \textbf{TX}      & \textbf{MT} & \textbf{Vocab} & \textbf{Cov (\%)} & \textbf{F1}   & \textbf{Prec} & \textbf{Rec} & \textbf{F1}   & \textbf{Prec} & \textbf{Rec}  \\\hline
OpenTag          & -&-              & 6,756  & 73.2    & 57.5 & 70.3 & 49.6   &  54.6 & 68.0   & 47.3   \\\hline
Title+id    & \checkmark &   -               & 6,400  & 69.1    & 55.7 & 70.6 & 46.9   & 53.3 & 68.9 & 45.1   \\
Title+name   & \checkmark & -& 5,328  & 70.6    & 56.9 & 71.2 & 48.4    & 54.2 & 69.1 & 46.3   \\
Title+path    & \checkmark &  -               & 4,608  & 64.6    & 54.3 & 72.0   & 44.6   & 51.9 & 69.1 & 43.2   \\
Concat-wemb-Euclidean  & \checkmark &- & 9,768 & 76.3 & 60.1 & 71.6 & 52.9 & 57.4 & 69.0 & 50.6 \\
Concat-wemb-Poincaré   & \checkmark &  -       & 8,684  & 74.3    & 60.6 & 73.4 & 52.7  & 57.7 & 70.2 & 50.6   \\
Concat-LSTM-Euclidean  & \checkmark &- & 9,255 & 75.9 & 60.1 & 71.9 & 52.8 & 57.5 & 69.4 & 50.6\\
Concat-LSTM-Poincaré  & \checkmark & -& 8,893 & 75.2 & 60.8 & 72.9 & 53.2  & 57.9& 70.3 & 50.9\\
Gate-Poincaré     & \checkmark &  -       & 9,690  & 77.1    & 60.6 & 71.5 & 53.5   & 57.7 & 69.3 & 51.0     \\
CondSelfAtt-Poincaré  & \checkmark & -       & 12,558 & 83.1    & 61.9 & 68.8 & 57.0   & 58.3 & 66.5 & 53.1   \\\hline
MT-flat             & - &      \checkmark      & 8,699  & 72.2    & 60.9 & 74.7 & 52.4  & 57.8 & 70.3 & 50.5   \\
MT-hier                & - &      \checkmark          & 9,528  & 73.4    & 61.5 & 74.5 & 53.2 & 58.3 & \textbf{70.9} & 51.1   \\\hline
Concat \& MT-hier    &  \checkmark &      \checkmark     & 9,316  & 74.6    & 62.3 & \textbf{75.0}   & 54.3 & 59.0   & 70.8 & 52.1   \\
Gate \& MT-hier  &  \checkmark &      \checkmark      & 10,845 & 80.0      & 61.1 & 70.7 & 54.8  & 57.9 & 67.9 & 51.8   \\
CondSelfAtt \& MT-hier (TXtract) &  \checkmark &      \checkmark & \textbf{13,093} & \textbf{83.9}    & \textbf{63.3} & 70.9 & \textbf{57.8}   & \textbf{59.3} & 68.4 & \textbf{53.8}\\ 

\hline 
\end{tabular}}
\caption{Results for \emph{flavor} extraction across all categories. ``TX'' column indicates whether the taxonomy is leveraged for attribute value extraction (Section~\ref{s:model-taxonomy-aware-extraction}). ``MT'' column indicates whether multi-task learning is used (Section~\ref{ss:multi-task-learning}).}
\label{tab:flavor-results}
\end{table*}

\begin{figure*}
    \centering
    \includegraphics[height = 9cm, width = 12cm]{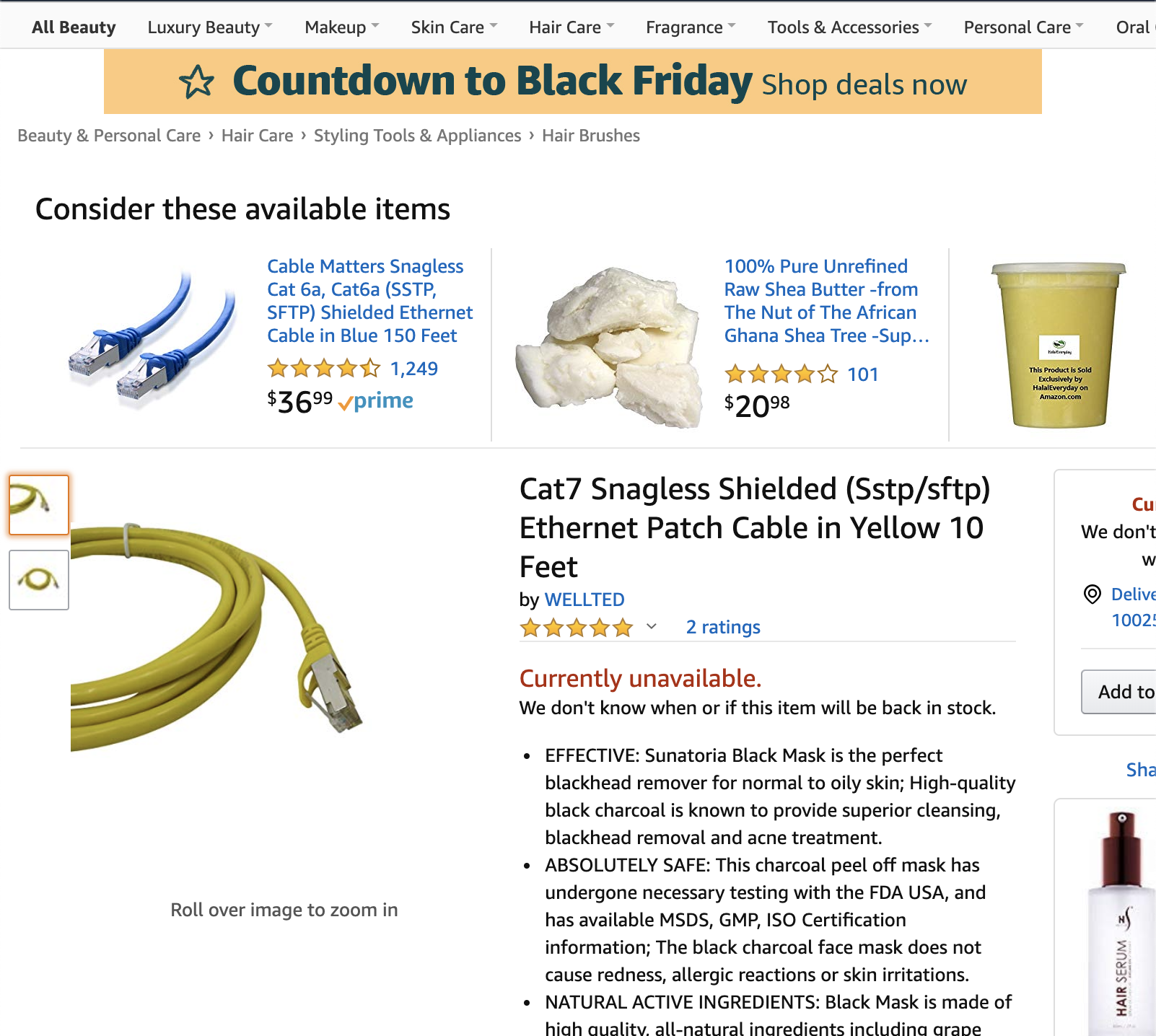}
    \caption{Snapshot of \href{https://www.amazon.com/dp/B012AE5EP4}{{https://www.amazon.com/dp/B012AE5EP4}}. This ethernet cable has been erroneously assigned under ``Hair Brushes'' category. (The assignment can be seen on the top left part of the screenshot.)}
    \label{fig:my_label}
\end{figure*}

\begin{figure*}
    \centering
    \includegraphics[height = 9cm, width = 12cm]{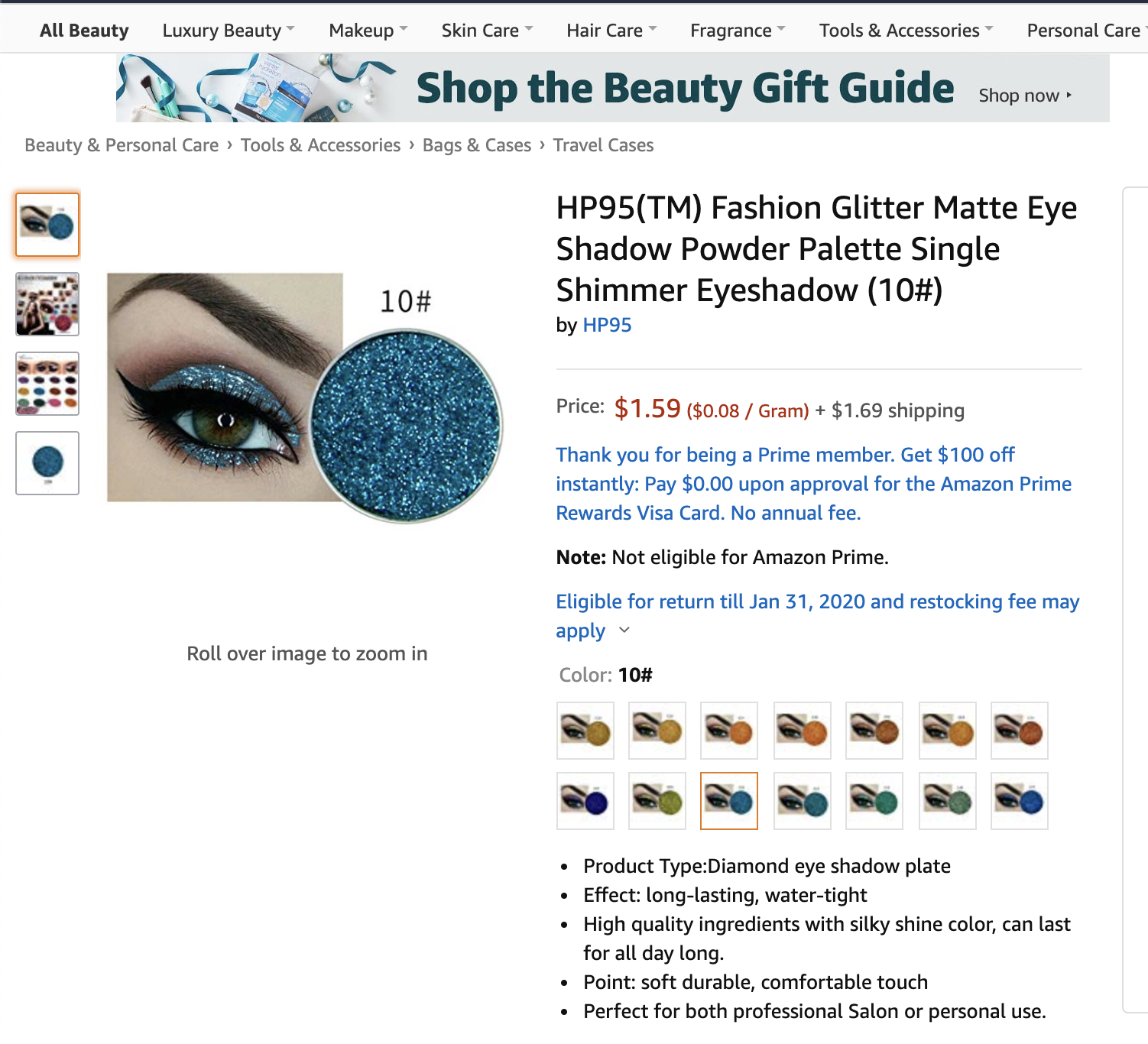}
    \caption{Snapshot of \href{https://www.amazon.com/dp/B07BBM5B33}{{https://www.amazon.com/dp/B07BBM5B33}}. This eye shadow product has been erroneously assigned under ``Travel Cases'' category. (The assignment can be seen on the top left part of the screenshot.)}
    \label{fig:my_label}
\end{figure*}
\end{document}